\documentclass{ieeeaccess}
\usepackage{cite}
\usepackage{amsmath,amssymb,amsfonts}
\usepackage{algorithmic}
\usepackage{graphicx}
\usepackage{textcomp}
\usepackage{array}

\usepackage{float} 
\usepackage{multirow}

\usepackage{bm}
\makeatletter
\AtBeginDocument{\DeclareMathVersion{bold}
\SetSymbolFont{operators}{bold}{T1}{times}{b}{n}
\SetSymbolFont{NewLetters}{bold}{T1}{times}{b}{it}
\SetMathAlphabet{\mathrm}{bold}{T1}{times}{b}{n}
\SetMathAlphabet{\mathit}{bold}{T1}{times}{b}{it}
\SetMathAlphabet{\mathbf}{bold}{T1}{times}{b}{n}
\SetMathAlphabet{\mathtt}{bold}{OT1}{pcr}{b}{n}
\SetSymbolFont{symbols}{bold}{OMS}{cmsy}{b}{n}
\renewcommand\boldmath{\@nomath\boldmath\mathversion{bold}}}
\makeatother

\def\BibTeX{{\rm B\kern-.05em{\sc i\kern-.025em b}\kern-.08em
    T\kern-.1667em\lower.7ex\hbox{E}\kern-.125emX}}

\begin{document}
\history{Date of publication xxxx 00, 0000, date of current version xxxx 00, 0000.}
\doi{10.1109/ACCESS.2024.0429000}

\title{AuxDepthNet: Real-Time Monocular 3D Object Detection with Depth-Sensitive Features}
\author{\uppercase{Ruochen Zhang}\authorrefmark{1},
        \uppercase{Hyeung-Sik Choi}\authorrefmark{1},
        \uppercase{Dongwook Jung}\authorrefmark{1}, 
        \uppercase{Phan Huy Nam Anh}\authorrefmark{1}, 
        \uppercase{Sang-Ki Jeong}\authorrefmark{2},
        \uppercase{Zihao Zhu}\authorrefmark{1}}

\address[1]{Department of Mechanical Engineering, National Korea Maritime and Ocean University, Busan 49112, Republic of Korea}
\address[2]{Maritime ICT and Mobility Research Department, Korea Institute of Ocean Science and Technology,
Busan 49111, Republic of Korea}


\markboth
{Author \headeretal: Preparation of Papers for IEEE TRANSACTIONS and JOURNALS}
{Author \headeretal: Preparation of Papers for IEEE TRANSACTIONS and JOURNALS}

\corresp{Corresponding author: Hyeung-Sik Choi (e-mail: hchoi@kmou.ac.kr).}

\begin{abstract}
Monocular 3D object detection is a challenging task in autonomous systems due to the lack of explicit depth information in single-view images. Existing methods often depend on external depth estimators or expensive sensors, which increase computational complexity and hinder real-time performance. To overcome these limitations, we propose AuxDepthNet, an efficient framework for real-time monocular 3D object detection that eliminates the reliance on external depth maps or pre-trained depth models. AuxDepthNet introduces two key components: the Auxiliary Depth Feature (ADF) module, which implicitly learns depth-sensitive features to improve spatial reasoning and computational efficiency, and the Depth Position Mapping (DPM) module, which embeds depth positional information directly into the detection process to enable accurate object localization and 3D bounding box regression. Leveraging the DepthFusion Transformer architecture, AuxDepthNet globally integrates visual and depth-sensitive features through depth-guided interactions, ensuring robust and efficient detection. Extensive experiments on the KITTI dataset show that AuxDepthNet achieves state-of-the-art performance, with \( \text{AP}_{3D} \) scores of 24.72\% (Easy), 18.63\% (Moderate), and 15.31\% (Hard), and \( \text{AP}_{\text{BEV}} \) scores of 34.11\% (Easy), 25.18\% (Moderate), and 21.90\% (Hard) at an IoU threshold of 0.7.
\end{abstract}


\titlepgskip=-21pt

\maketitle

\section{Introduction}
\label{sec:introduction}
\PARstart{T}{hree-dimensional} (3D) object detection~\cite{he2020structure,lang2019pointpillars,shi2019pointrcnn,li2019stereo,wang2019pseudo} plays a critical role in autonomous driving and robotic perception by enabling machines to perceive and interact with their surroundings in a spatially aware manner. Traditionally, this task has relied on precise depth information from multiple sensors, such as LiDAR, stereo cameras, or depth sensors. While these systems provide high accuracy, they are often expensive and complex to deploy. In recent years, monocular 3D object detection has gained attention as a cost-effective alternative, requiring only a single RGB camera~\cite{brazil2020kinematic,chen2020monopair,ku2019monocular}. Despite its simplicity and accessibility, monocular 3D detection faces challenges due to the absence of explicit depth cues, making effective depth reasoning essential for improving accuracy and practicality.

\begin{figure}[t]
    \centering
    \includegraphics[width=1\linewidth]{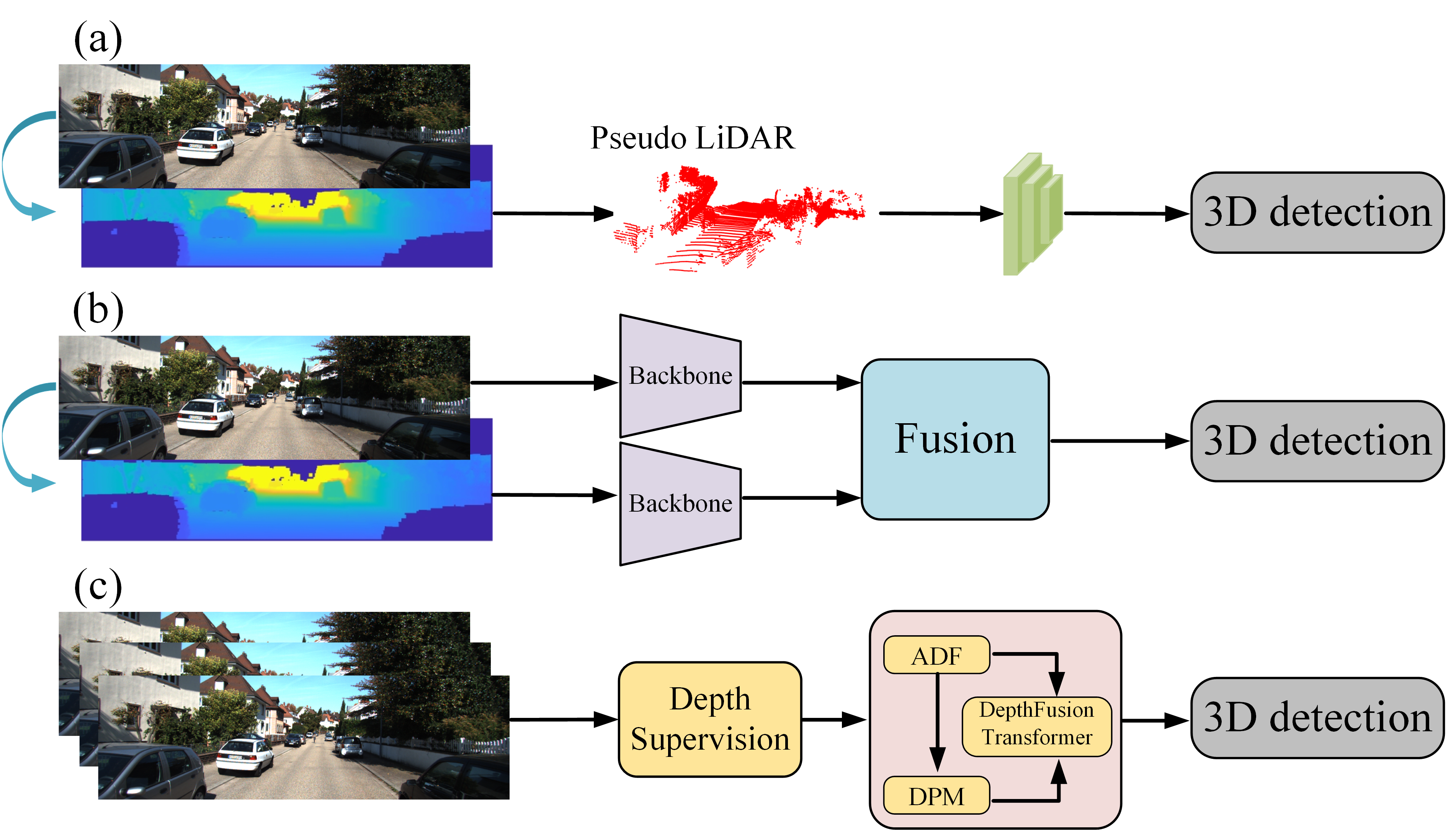}
    \caption{Representative Depth-Assisted Monocular 3D Object Detection Frameworks. (a) Depth estimation methods use monocular inputs to construct pseudo-LiDAR data, enabling LiDAR-style 3D detectors~\cite{ma2019accurate,wang2019pseudo,weng2019monocular}. (b) Fusion-based methods combine visual and depth features to improve object detection accuracy~\cite{ding2020learning,ouyang2020dynamic,wang2021depth}. (c) Our AuxDepthNet leverages depth guidance during training to develop depth-sensitive features and performs end-to-end 3D object detection without requiring external depth estimators.}
    \label{fig:Mono}
\end{figure}

Monocular 3D object detection methods can be broadly categorized into two approaches, as illustrated in Fig.~\ref{fig:Mono}. The first category utilizes pre-trained depth estimation models to generate pseudo-depth maps, which are combined with LiDAR-based 3D detectors for object recognition and localization~\cite{shen2023git, shen2023pbsl}. This approach, as exemplified by methods like Pseudo-LiDAR and F-Pointnet~\cite{ma2019accurate,wang2019pseudo,weng2019monocular}, achieves improved localization but suffers from inaccuracies in depth priors and high computational costs due to the reliance on external depth estimators, limiting their applicability in real-time scenarios.

The second category focuses on feature fusion, where features from monocular images and estimated depth maps are extracted and combined to enhance detection. Methods like D$^4$LCN and CaDDN~\cite{ding2020learning,ouyang2020dynamic}, shown in Fig.~\ref{fig:Mono}(b), use specialized convolutional architectures to integrate visual and depth features. While these methods demonstrate promising results, their reliance on the quality of depth maps and the complexity of their architectures can hinder efficiency and robustness in dynamic environments.

To address these challenges, we propose AuxDepthNet, a novel framework for real-time monocular 3D object detection that avoids the need for external depth estimators or pre-generated depth maps. AuxDepthNet introduces two key modules: the Auxiliary Depth Feature Module (ADF) and the Depth Position Mapping Module (DPM), as depicted in Fig.~\ref{fig:Mono}(c). These modules enable implicit learning of depth-sensitive features and inject depth positional cues directly into the detection pipeline. Conventional CNN-based monocular 3D detection frameworks often struggle to capture global dependencies and integrate multi-source depth features. In contrast, Transformers leverage self-attention mechanisms to process image and depth information simultaneously~\cite{kim2021hotr,zhu2020deformable}, allowing for end-to-end 3D object detection within a streamlined pipeline~\cite{carion2020end}. Built upon the DepthFusion Transformer architecture, our method efficiently integrates contextual and depth-sensitive features, achieving superior detection performance with reduced computational costs~\cite{weng2019monocular,ma2019accurate}.

The main contributions of this paper are as follows:
\begin{enumerate}
    \item We propose AuxDepthNet, a novel framework for efficient, real-time monocular 3D object detection that eliminates reliance on external depth maps or estimators.
    \item We design the Auxiliary Depth Feature Module (ADF) and Depth Position Mapping Module (DPM) to implicitly learn depth-sensitive features and encode depth positional cues into the detection process.
    \item We provide a plug-and-play architecture that can be seamlessly integrated into other image-based detection frameworks to enhance their 3D reasoning capabilities.
\end{enumerate}

\section{Related Work}\label{sec:rw}

Monocular 3D object detection has gained significant attention for its cost-effectiveness compared to sensor-based approaches, such as LiDAR and stereo cameras. Existing methods can be broadly categorized into depth-based methods and Transformer-based methods, which are discussed below.

\subsection{Monocular 3D Object Detection Methods}
Image-based 3D object detection leverages monocular or stereo images to estimate depth and generate 3D proposals. Approaches like MONO3D~\cite{ding2020learning}, MLF~\cite{xu2018multi}, and Frustum PointNet~\cite{qi2018frustum} use depth maps or disparity estimates to compute 3D coordinates, often integrating them into 2D detection pipelines. However, poor depth representations in these methods limit the ability of convolutional networks to accurately localize objects, especially at greater distances, leading to performance gaps compared to LiDAR-based systems.
Depth-based methods aim to bridge this gap by explicitly or implicitly leveraging depth information. One common approach is to estimate pseudo-depth maps from monocular images and utilize them for 3D object detection. For instance, Pseudo-LiDAR~\cite{wang2019pseudo} and F-PointNet~\cite{weng2019monocular} rely on pre-trained depth models to generate point clouds, which are then processed with LiDAR-style 3D detectors. While these methods improve spatial reasoning, they suffer from depth estimation errors and increased computational costs due to reliance on external depth estimators.

Another line of work focuses on integrating monocular image features with estimated depth information in a feature fusion framework. Methods like D$^4$LCN~\cite{ding2020learning} and CaDDN~\cite{reading2021categorical} incorporate depth-sensitive features into the 2D detection pipeline to enhance accuracy. However, these approaches are constrained by the quality of the estimated depth maps and often rely on complex architectures, limiting their real-time applicability.
In contrast, our proposed method avoids reliance on external depth estimators by directly learning depth-sensitive features through the Auxiliary Depth Feature (ADF) module. This design allows AuxDepthNet to achieve high accuracy while maintaining computational efficiency.

\subsection{Transformer in Monocular 3D Object Detection}
The recent success of Transformers in computer vision has inspired their application in monocular 3D object detection. Unlike convolutional neural networks (CNNs), Transformers capture global spatial dependencies using self-attention mechanisms, addressing the limitations of local feature extraction. For example, MonoDETR~\cite{zhang2023monodetr} employs a dual-encoder architecture and a depth-guided decoder to improve depth representation and object localization. Similarly, MonoPSTR~\cite{yang2024monopstr} introduces scale-aware attention and position-coded queries to enhance detection precision and efficiency.
Recently, diffusion models~\cite{shen2023advancing,shen2024boosting,shen2024imagdressing,shen2024imagpose} have also demonstrated their potential in depth estimation by iteratively refining depth predictions, offering an alternative to traditional depth map generation methods. These models could complement Transformer-based architectures by providing robust depth priors for enhanced detection.

Despite their advancements, Transformer-based methods often rely on pre-computed depth maps or handcrafted priors to guide detection~\cite{carion2020end,zou2021end}. This dependence on external inputs can hinder adaptability to diverse scenarios. AuxDepthNet addresses this limitation by embedding depth reasoning directly into the network via the Auxiliary Depth Feature (ADF) and Depth Position Mapping (DPM) modules. These modules enable the model to implicitly learn depth-sensitive features, eliminating external dependencies and delivering robust, scalable performance.

\begin{figure*}[t]
    \centering
    \includegraphics[width=1.033\textwidth]{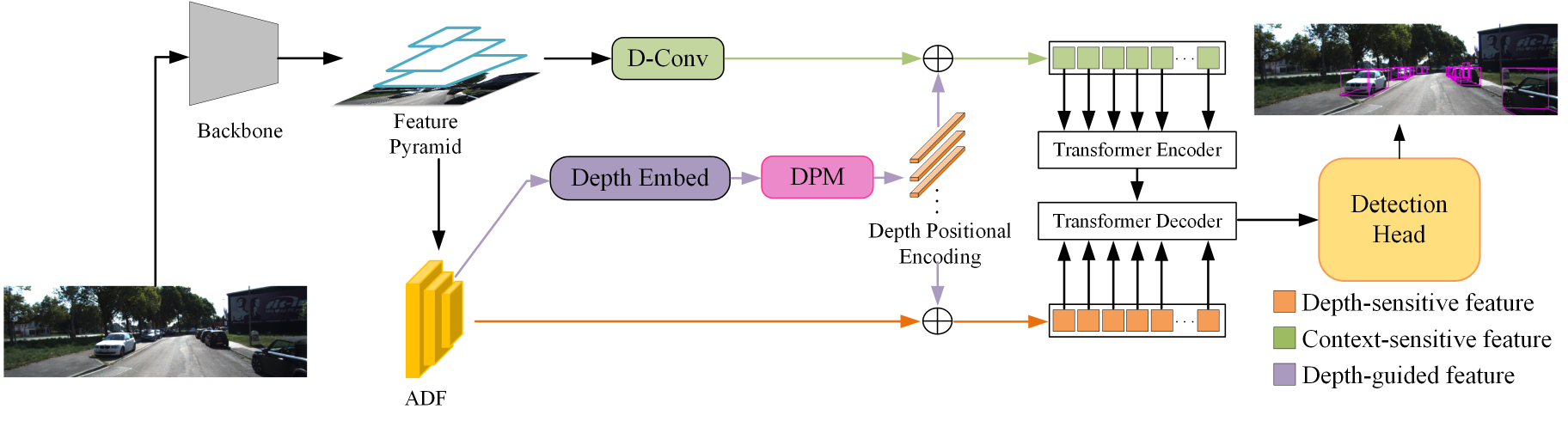}
       \caption{
\textbf{The overall framework of our proposed method.}
AuxDepthNet enhances monocular 3D object detection by integrating depth-sensitive, context-sensitive, and depth-guided features. The Auxiliary Depth Feature (ADF) module encodes depth-related cues without pre-computed depth maps. Context-sensitive features are refined by a feature pyramid and DepthFusion Transformer (DFT), providing semantic and spatial context. The Depth Position Mapping (DPM) module embeds depth-based positional information for precise 3D localization. This integration captures local and global spatial relationships efficiently, delivering robust 2D and 3D detection.}
    \label{fig:mono2}
\end{figure*}

\section{Proposed Method}\label{sec:method} 
\subsection{Overview}
As shown in Fig.\ref{fig:mono2}, the AuxDepthNet architecture leverages a multi-faceted feature representation to enhance monocular 3D object detection. Specifically, the model incorporates depth-sensitive features, context-sensitive features, and depth-guided features to capture and integrate crucial information at different stages. The depth-sensitive features are extracted through the Auxiliary Depth Feature (ADF) module, which implicitly encodes depth-related cues via auxiliary learning, eliminating the need for pre-computed depth maps. Context-sensitive features, generated by the backbone and refined through the feature pyramid and DepthFusion Transformer (DFT), provide semantic and spatial context for accurate object detection. Depth-guided features, enhanced by the Depth Position Mapping (DPM) module and positional encoding, embed depth-positional cues into the feature space, enabling precise spatial reasoning and robust 3D localization. The combined integration of these feature types ensures the framework captures both local and global spatial relationships, providing robust 2D and 3D object detection with minimal computational overhead.

\subsection{Depth-Sensitive Feature Enhancement}
Current depth-assisted approaches face challenges in generalizing to varied datasets and environments, which hampers their adaptability. Moreover, their dependency on external depth sensors or estimators not only increases hardware requirements but also risks propagating inaccuracies into the detection pipeline~\cite{ding2020learning,wang2021depth,wang2019pseudo,carion2020end}. To address these challenges, we propose the Auxiliary Depth Feature (ADF) module, which leverages auxiliary supervision during training to implicitly learn depth-sensitive features, eliminating the need for external depth maps or estimators.

The ADF module is designed as a lightweight and efficient solution that enhances depth reasoning and spatial localization while maintaining low computational cost. Unlike previous methods~\cite{shi2019pointrcnn,wang2019pseudo,ku2019monocular}, which rely on pre-computed depth maps or external estimators, the ADF module captures depth-sensitive features directly from the input feature map using auxiliary learning. This approach ensures scalability, improves generalization across datasets, and enables real-time applicability.

\begin{figure}[t]
    \centering
    \includegraphics[width=1\linewidth]{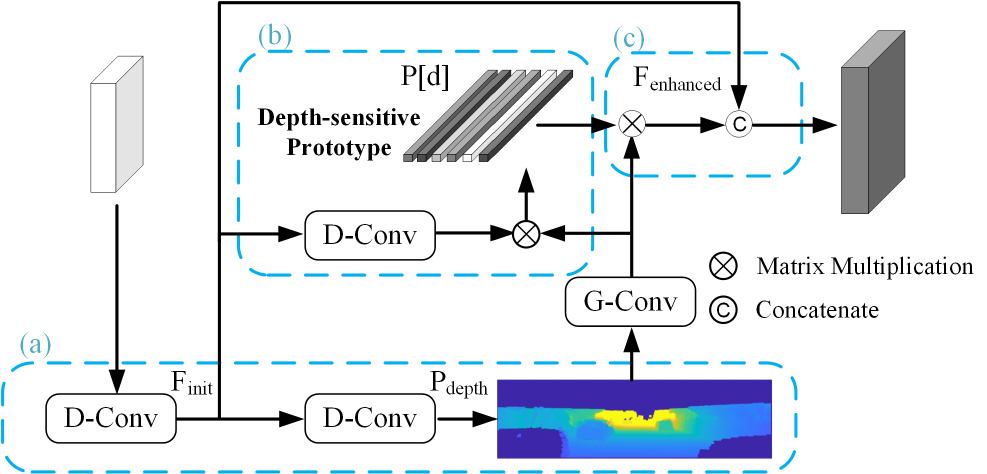}
\caption{
\textbf{
Architecture of the Auxiliary Depth Feature Module
}
(a)Generate the initial depth-sensitive feature \(F_{\text{init}}\) and determine the depth distribution \(P_{\text{depth}}\).
(b) P[d] represents the feature representation of the depth prototype.
(c) The depth prototype enhanced feature \(F_{\text{enhanced}}\) is generated and fused with \(F_{\text{init}}\).
}
\label{fig:Mono5}
\end{figure}

As illustrated in Fig.~\ref{fig:Mono5}, the Auxiliary Depth Feature (ADF) module operates in three stages to enhance depth-sensitive features. First, it generates initial depth-sensitive features through an auxiliary supervision task, which predicts a probability distribution over discretized depth bins for each pixel. Next, the module learns depth prototypes by leveraging spatial-depth attention mechanisms, aggregating features across spatial dimensions to capture meaningful depth relationships. Finally, these depth prototypes are projected back to refine the feature map, producing an enhanced representation that integrates depth-sensitive information and spatial context, thereby improving the accuracy and robustness of 3D object detection.

The following sections detail each step of the ADF module and its role in improving depth representation for 3D object detection.

\subsubsection{Extracting Foundational depth-sensitive Features.}
In the Auxiliary Depth Feature (ADF) module, foundational depth-sensitive features are generated using an auxiliary depth estimation task modeled as a sequential classification problem. Given the input feature map $F \in \mathbb{R}^{C \times H \times W}$from the backbone, we apply two convolutional layers to predict the probability distribution of discretized depth bins $ D_b \in \mathbb{R}^{D \times H \times W}$, where $D_b$ denotes the number of depth categories. The predicted probability represents the confidence of each pixel belonging to a specific depth bin:
\begin{align}
P_{i, j, d} &= \frac{\exp \left(z_{i, j, d}\right)}{\sum_{k=1}^{D_b} \exp \left(z_{i, j, k}\right)}.
\end{align}
where $z_{i, j, d}$ is the raw score for pixel $(i, j)$ in bin $n$.

To discretize continuous depth values into bins, we utilize Linear-Increasing Discretization (LID), which refines depth granularity for closer objects while allocating broader intervals for farther ones. The discretization can be formulated as:
\begin{align}
\mathit{B}_{i}=\left\{\begin{array}{ll}\frac{(i+1)^{2}}{D}, & \text { if } i<\sqrt{D}, \\\frac{(i+1)}{D}, & \text { otherwise. }\end{array}\right.
\end{align}

By applying depthwise separable convolutions, the intermediate feature map $\mathbf{X} \in \mathbb{R}^{C \times H \times W}$
  efficiently captures depth-sensitive features with reduced computational overhead. These features, further enhanced through attention mechanisms, provide a robust foundation for downstream 3D object detection tasks.

\subsubsection{Depth-Sensitive Prototype Representation module.} This module aims to refine feature representations through depth prototype learning and enhancement. Starting from the initial depth-sensitive feature map \( F_{\text{init}} \in \mathbb{R}^{C \times H \times W} \), generated by the backbone network, the module predicts the depth distribution \( P_{\text{depth}} \in \mathbb{R}^{D \times H \times W} \) for each pixel, where \( C \) is the feature dimension, \( H \) and \( W \) are spatial dimensions, and \( D \) represents the number of depth bins. The predicted probability represents the confidence of each pixel belonging to a specific depth bin, which is computed using softmax normalization. To enhance the feature extraction process, dilated convolutions(D-Conv) are employed to enlarge the receptive field while maintaining spatial resolution. Using the predicted depth distribution, the module estimates depth prototype representations \( P[d] \in \mathbb{R}^{C} \) for each depth bin \( d \) by aggregating features across all pixels:
\begin{align}
\mathit{P}[d] 
= \sum_{i=1}^{N} \operatorname{Attention}(i, d) \times \mathit{F}_{\mathit{init}}[i], 
\quad d \in \{1, \ldots, D\}.
\end{align}
where $\mathit{P}[d]$, \( N = H \times W \) represents the total number of spatial positions, and \( \text{Attention}(i, d) \) denotes the attention weight for pixel \( i \) and depth bin \( d \), derived from the similarity between \( \mathit{F}_{\text{init}} \) and \( \mathit{P}_{\text{depth}} \).

\subsubsection{Feature enhancement with depth prototype.} 
The depth prototypes are then projected back to enhance the feature map, creating an updated representation:
\begin{align}
\mathit{F}_{\mathit{enhanced}}[i]
= \sum_{d=1}^{D} \operatorname{Attention}(i, d) \times \mathit{P}[d],
\quad i \in \{1, \ldots, N\}.
\end{align}

Finally, the enhanced feature map $\mathit{F}_{\text{enhanced}} \in \mathbb{R}^{C \times H \times W}$ undergoes a convolutional refinement to integrate depth-sensitive
information.As a result, the enhanced depth feature is generated by combining the initial  depth-sensitive feature with the reconstructed features through concatenation This mechanism ensures that the depth-sensitive prototypes effectively capture spatial-depth relationships, significantly improving feature quality for 3D object detection tasks.

\subsection{Extracting Foundational Depth-Sensitive Features}
Inspired by the success of Transformer architectures in capturing long-range dependencies and modeling global relationships~\cite{vaswani2017attention}, we introduce the Depth Position Mapping (DPM) module as a key component of our framework. The DPM module is designed to address the challenges of integrating spatial and depth cues in monocular 3D object detection. Unlike traditional methods that rely solely on local visual features or predefined depth priors, DPM leverages a depth-guided mapping mechanism to explicitly encode positional depth information into the feature space. By embedding depth positions into learnable queries within an encoder-decoder structure, the module enables precise alignment of spatial and depth representations. This design enhances the understanding of scene-level depth geometry while improving the quality of feature fusion for downstream 3D attribute prediction. The DPM module thus provides a robust and efficient solution for bridging the gap between depth estimation and object localization in monocular 3D object detection.

\subsubsection{Transformer Encoder.} The Transformer Encoder in the Depth Position Mapping (DPM) module plays a critical role in capturing global dependencies and refining feature representations by leveraging self-attention mechanisms. It enables the model to incorporate contextual information across the entire sequence, facilitating a deeper understanding of spatial and depth relationships. The encoder refines input features by employing a multi-head self-attention mechanism and a feed-forward neural network (FFN)~\cite{carion2020end,zou2021end}.Given the input feature tensor $\mathbf{X} \in \mathbb{R}^{N \times L \times C}$ where $N$ is the batch size, $L$ is the sequence length, and $C$ is the feature dimensionality, the encoder projects X into query ($\mathit{Q}_{\text{f}}$), key ($\mathit{K}_{\text{f}}$), and value ($\mathit{V}_{\text{f}}$) matrices, where $\mathit{Q}_{\text{f}}, \mathit{K}_{\text{f}}, \mathit{V}_{\text{f}} \in \mathbb{R}^{N \times L \times C}$. These are further divided into $H$ attention heads, with each head having a dimensionality of $D = C / H$. The attention weights are computed as:

\begin{align}
\operatorname{Attention}\left(\mathit{Q}_{\text{f}}, \mathit{K}_{\text{f}}\right)=\operatorname{Softmax}\left(\frac{\mathit{Q}_{\text{f}} \times \mathit{K}_{\text{f}}^{\top}}{\sqrt{D}}\right).
\end{align}

enabling the encoder to capture long-range dependencies across the sequence. A linear attention mechanism enhances computational efficiency by transforming~\cite{vaswani2017attention} $\mathit{Q}_{\text {f}}$ and $\mathit{K}_{\text {f}}$ , using an ELU-based feature map. The output of the attention layer, $\mathbf{Z} \in \mathbb{R}^{N \times L \times C}$ , is combined with the input via a residual connection and normalized using layer normalization, $\mathbf{X}_{1}=\operatorname{Norm}(\mathbf{X}+\mathbf{Z})$ . This is followed by a two-layer FFN, $\operatorname{FFN}(\mathit{X})=\operatorname{ReLU}\left(\mathit{X} \times \mathit{W}_{1}\right) \times \mathit{W}_{2}$
, where ${W}_{1}$ and ${W}_{2}$ are learnable weight matrices. The FFN output is also added back to the input with a residual connection and normalized, $\mathbf{X}_{\text {out }}=\operatorname{Norm}\left(\mathbf{X}_{1}+\operatorname{FFN}\left(\mathbf{X}_{1}\right)\right)$. The final output, $\mathbf{X}_{\text {out }} \in \mathbb{R}^{N \times L \times C}$ , integrates depth-sensitive global context and long-range dependencies, providing refined features for downstream tasks.

\subsubsection{Transformer Decoder.}
The Transformer Decoder is designed to integrate depth-sensitive features with contextual information, enabling robust representation refinement for downstream tasks~\cite{vaswani2017attention}. By employing a multi-layer structure, each layer consists of self-attention, cross-attention, and a feed-forward network. Self-attention operates on the input queries to capture internal dependencies, while cross-attention aligns these queries with encoder outputs, ensuring the effective fusion of spatial and contextual features. Specifically, cross-attention uses depth-sensitive features as queries and leverages contextual embeddings as keys and values, allowing the decoder to focus on relevant spatial-depth relationships. Each layer applies residual connections and normalization to stabilize feature updates, while a feed-forward network further enhances the refined representations. Through iterative processing across multiple layers, the decoder progressively aligns and integrates depth-sensitive features and contextual cues, resulting in task-specific predictions with enhanced accuracy and robustness.
\begin{figure}[t]
    \centering
    \includegraphics[width=1\linewidth]{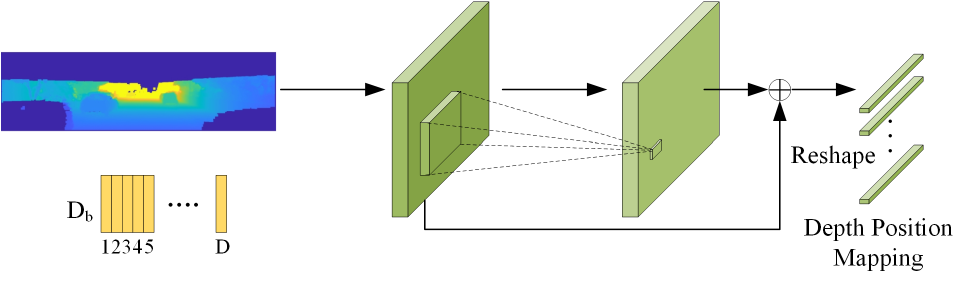}
\caption{
\textbf{Overview of the proposed Depth Position Mapping (DPM) module.}
This process aligns spatial features with depth information, enhancing the model's 3D geometric understanding.}
\label{fig:Mono6}
\end{figure}

\subsubsection{Depth Position Mapping (DPM) module.} The Depth Position Mapping (DPM) module embeds depth-related positional information to enhance the model’s understanding of spatial-depth relationships as shown in Fig \ref{fig:Mono6}. Given the input features $\mathbf{X} \in \mathbb{R}^{B \times N \times C}$, where $B$ is the batch size, $N=H \times W$ is the number of spatial positions, and $C$ is the number of spatial positions, and $\mathit{F} \in \mathbb{R}^{B \times C \times H \times W}$ . Based on previously predicted depth bins $\mathit{D}_{\mathit{b}}=\left[\mathit{d}_{1}, \ldots, \mathit{d}_{D}\right] \in \mathbb{R}^{D \times C}$ , where $D$ is the number of depth bins and $\mathbf{d}_{i}$ represents the learnable embedding for the $i$-th depth category, the depth features are locally refined using a depthwise separable convolution with a $3 \times 3$ kernel. The operation is defined as:
\begin{align}
\mathit{F}^{\prime} & = \operatorname{Conv}_{3 \times 3}(\mathit{F})+\mathit{F}.
\end{align}
where $\operatorname{Conv}_{3 \times 3}$ integrates local depth information while maintaining computational efficiency. The resulting depth-sensitive feature map $\mathit{F}^{\prime}$ is then flattened back to $\mathit{R}^{B \times N \times C}$ ensuring alignment between the depth positional mappings and the spatial features. By embedding depth-sensitive positional cues through $\mathit{D}_{\mathrm{b}}$, the DPM module improves the model’s ability to capture 3D geometric structures, while preserving efficiency and scalability for downstream tasks.

\subsection{Loss Function}  
The proposed framework adopts a single-stage detector design, which directly predicts object bounding boxes and class probabilities using pre-defined 2D-3D anchors. This architecture is optimized for efficient object detection and depth estimation in a single forward pass. To handle challenges such as class imbalance and bounding box regression, the framework incorporates Focal Loss [1] for classification and Smooth L1 Loss [2] for bounding box regression. These loss functions are combined with a custom depth loss to ensure robust performance across all tasks.

\setlength{\tabcolsep}{3pt} 
\begin{table*}[!htbp]
\caption{\textbf{Detection Performance of Car Category on the KITTI Test Set}}
\label{tab:kitti_test_car}
\renewcommand{\arraystretch}{1.2} 
\centering
\noindent\makebox[\textwidth]{
\begin{tabular}{|>{\centering\arraybackslash}p{115pt}|>{\centering\arraybackslash}p{55pt}|
                >{\centering\arraybackslash}p{40pt}|>{\centering\arraybackslash}p{40pt}|>{\centering\arraybackslash}p{40pt}|
                >{\centering\arraybackslash}p{40pt}|>{\centering\arraybackslash}p{40pt}|>{\centering\arraybackslash}p{40pt}|}
\hline
\textbf{Method} & \textbf{Reference} & \multicolumn{3}{c|}{\apthreeD@IoU=0.7} & \multicolumn{3}{c|}{\apBev@IoU=0.7} \\ 
\cline{3-8}
 & & Easy & Mod. & Hard & Easy & Mod. & Hard \\ 
\hline
DDMP-3D~\cite{wang2021depth} & CVPR 2021 & 19.71 & 12.78 & 9.80 & 28.08 & 17.89 & 13.44 \\ 
CaDDN~\cite{reading2021categorical} & CVPR 2021 & 19.17 & 13.41 & 11.46 & 27.94 & 18.91 & 17.19 \\
DFRNet~\cite{zou2021devil} & ICCV 2021 & 19.40 & 13.63 & 10.35 & 28.17 & 19.17 & 14.84 \\
MonoEF~\cite{zhou2021monocular} & CVPR 2021 & 21.29 & 13.87 & 11.71 & 29.03 & 19.70 & 17.26 \\
MonoFlex~\cite{zhang2021objects} & CVPR 2021 & 19.94 & 13.89 & 12.07 & 28.23 & 19.75 & 16.89 \\ 
GUPNet~\cite{lu2021geometry} & ICCV 2021 & 20.11 & 14.20 & 11.77 & - & - & - \\ 
MonoDTR~\cite{huang2022monodtr} & CVPR 2022 & 21.99 & 15.39 & 12.73 & 28.59 & 20.38 & 17.14 \\
MonoJSG~\cite{lian2022monojsg} & CVPR 2022 & 24.69 & 16.14 & 13.64 & 32.59 & 21.26 & 18.18 \\
DEVIANT~\cite{kumar2022deviant} & ECCV 2022 & 21.88 & 14.46 & 11.89 & 29.65 & 20.44 & 17.43 \\
DCD~\cite{li2022densely} & ECCV 2022 & 23.94 & 17.38 & \textcolor{blue}{15.32} & 32.55 & 21.50 & 18.25 \\
DID-M3D~\cite{peng2022did} & ECCV 2022 & 24.40 & 16.29 & 13.75 & 32.95 & 22.76 & 19.83 \\
Cube R-CNN~\cite{brazil2023omni3d} & CVPR 2023 & 23.59 & 15.01 & 12.56 & 31.70 & 21.20 & 18.43 \\
MonoUNI~\cite{jinrang2024monouni} & NeurIPS 2023 & \textcolor{red}{24.75} & 16.73 & 13.49 & 33.28 & 23.05 & 19.39 \\ 
MonoATT~\cite{zhou2023monoatt} & CVPR 2023 & \textcolor{blue}{24.72} & \textcolor{blue}{17.37} & 15.00 & \textcolor{red}{36.87} & \textcolor{blue}{24.42} & \textcolor{blue}{21.88} \\ 
AuxDepthNet (Ours) & - & \textcolor{blue}{24.72} & \textcolor{red}{18.63} & \textcolor{red}{15.31} & \textcolor{blue}{34.11} & \textcolor{red}{25.18} & \textcolor{red}{21.90} \\ 
\hline
\end{tabular}}
\vspace{2mm}
\raggedright \small{The best and second-best results are highlighted in \textcolor{red}{red} and \textcolor{blue}{blue}, respectively.}
\end{table*}
For 3D object detection, the classification loss $\mathcal{L}_{\mathrm{cls}}$ and regression loss $\mathcal{L}_{\mathrm{reg}}$ are defined for positive anchors. The Focal Loss for classification is used to mitigate the impact of class imbalance by down-weighting easy samples:
\begin{align}
\mathcal{L}_{\mathrm{cls}}=\frac{1}{N_{\mathrm{pos}}} \sum_{i=1}^{N} \operatorname{FocalLoss}\left(\mathit{p}_{i}, \mathit{p}_{i}^{\mathrm{gt}}\right).
\end{align}
where $N_{\text {pos }}$ is the number of positive anchors. The Smooth L1 Loss for bounding box regression minimizes the deviation between predicted and ground truth box parameters:
\begin{align}
\mathcal{L}_{\mathrm{reg}}=\frac{1}{N_{\mathrm{pos}}} \sum_{i=1}^{N_{\mathrm{pos}}} \operatorname{SmoothL1}\left(\mathit{b}_{i}, \mathit{b}_{i}^{\mathrm{gt}}\right).
\end{align}
Depth estimation is treated as a discrete classification problem, where depth ground truth values are projected into bins using Linear Increasing Discretization (LID)~\cite{reading2021categorical,tang2020center3d}. A focal-style depth loss is applied to prioritize confident predictions:
\begin{align}
\mathcal{L}_{\text {depth }}=-\frac{1}{N} \sum_{i=1}^{N} \mathit{w}_{i} \times \mathit{d}_{i}^{\mathrm{gt}} \times \log \left(\mathit{d}_{i}\right).
\end{align}
where $\mathbf{w}_{i}$ adjusts the contribution of each depth bin based on its confidence. The total loss is a weighted combination of these components:
\begin{align}
\mathcal{L}_{\text {total }}=\mathcal{L}_{\text {cls }}+\lambda_{\text {reg }} \times \mathcal{L}_{\text {reg }}+\lambda_{\text {depth }} \times \mathcal{L}_{\text {depth }}.
\end{align}

where $\lambda_{\text {reg }}$ and $\lambda_{\text {depth }}$ control the weights of the regression and depth losses, respectively.During training, anchors with an intersection-over-union (IoU) greater than 0.5 with the ground truth boxes are selected for optimization. By combining classification, regression, and depth estimation objectives, the proposed loss function ensures a balanced and effective training process. The use of Focal Loss and Smooth L1 Loss further enhances the robustness of the framework in handling challenging 3D object detection scenarios.

\section{Experiment and Analysis}\label{sec:exp}  
\subsection{Dataset}
The KITTI dataset is a widely used benchmark for autonomous driving research, providing labeled data for tasks like 3D object detection, stereo vision, and SLAM. It includes 7,481 training images and 7,518 test images, captured with stereo cameras and LiDAR in real-world driving scenarios. KITTI evaluates 3D object detection across three categories (Car, Pedestrian, Cyclist) and three difficulty levels (Easy, Moderate, Hard), making it a standard for testing perception algorithms.

\subsection{Evaluation Metrics}  
The KITTI dataset evaluates the performance of 3D object detection and bird’s-eye view (BEV) detection using Average Precision (AP) as the primary metric. To ensure a more accurate evaluation, it adopts the $\mathrm{AP}_{40}$ metric, which calculates the average precision across 40 evenly spaced recall positions, reducing potential bias from fewer recall points in the original AP metric.

Detection tasks are divided into three difficulty levels—Easy, Moderate, and Hard—based on object attributes such as occlusion, truncation, and size. The dataset includes three object categories: Car, Pedestrian, and Cyclist. The evaluation employs intersection over union (IoU) thresholds of 0.7 for the Car category and 0.5 for Pedestrian and Cyclist.

\subsection{Implementation Details} 
The proposed AuxDepthNet framework was evaluated on the KITTI dataset, targeting the Car category. Input images were resized to a resolution of $1280 \times 288$ The model was trained for 120 epochs with a batch size of 16, using the Adam optimizer with an initial learning rate of $1 \times
10^{-4}$ and no weight decay. A cosine annealing learning rate scheduler was employed, with the minimum learning rate set to $5 \times 10^{-6}$.

To enhance the robustness of the model, several data augmentation techniques were applied during training. These included photometric distortions (e.g., random adjustments to brightness, contrast, saturation, and hue), cropping the top 100 pixels of the image, random horizontal flipping with a probability of 0.5, resizing to the target resolution, and normalization using the RGB mean values $[0.485,0.456,0.406]$  and standard deviations $[0.229,0.224,0.225]$.

For testing, augmentation was limited to cropping, resizing, and normalization without photometric distortions or flipping.The detection head employed an anchor-based design, with IoU thresholds of 0.5 for positive samples and 0.4 for negative samples. The loss function combined focal loss with smooth L1 loss to balance the classification and regression tasks. During inference, non-maximum suppression (NMS) was applied with an IoU threshold of 0.4, and predictions with a confidence score below 0.75 were discarded.

\subsection{Comparison with State-of-the-art Methods} 
\subsubsection{KITTI Benchmark}
To evaluate the effectiveness of AuxDepthNet, we compare its performance with state-of-the-art monocular 3D object detection methods on the KITTI 3D Object Detection Benchmark as shown in Table \ref{tab:kitti_test_car}. The benchmark measures Average Precision (AP) under IoU 0.7 for both 3D detection (\( \text{AP}_{3D} \)) and bird's-eye view detection (\( \text{AP}_{\text{BEV}} \)) across three levels: Easy, Moderate, and Hard.
From Table 1, AuxDepthNet achieves competitive results. For 3D detection, it obtains an AP of 24.72, 18.63, and 15.31 on Easy, Moderate, and Hard levels, respectively. Compared to MonoUNI (NeurIPS 2023) with 24.75, 16.73, and 13.43, our method improves the Moderate and Hard scores by 1.90 and 1.88 points. Similarly, AuxDepthNet outperforms Cube R-CNN (CVPR 2023), which achieves 23.59, 15.01, and 12.65, across all three categories.
For BEV detection, AuxDepthNet achieves an AP of 34.11, 25.18, and 21.90. Compared to MonoATT (CVPR 2023), which achieves 36.87, 24.42, and 21.88, our method improves the Moderate score by 0.76 points while maintaining competitive performance on other levels.
Overall, AuxDepthNet surpasses MonoUNI, Cube R-CNN, and MonoATT, especially on Moderate and Hard levels. These results demonstrate the effectiveness of our Depth Position Mapping (DPM) module in capturing depth-sensitive features and enhancing 3D detection accuracy on the KITTI benchmark.

\subsection{Ablation Studies and Analysis}  
\subsubsection{Effectiveness of each proposed components.}
To assess the impact of each component in the AuxDepthNet framework, we conducted ablation studies highlighting their contributions to overall performance.(a)The baseline utilizes only context-aware features for 3D object detection, without incorporating depth-sensitive modules.(b)Context-aware and depth-sensitive features were integrated using a convolutional concatenate operation as an alternative to the depth-sensitive transformer module.(c)Depth-sensitive features in the transformer were replaced with object queries similar to DETR, creating a baseline with a DETR-like transformer. (d)Depth-sensitive features were replaced with those extracted from depth images generated by a pretrained depth estimator (DORN), instead of being learned in an end-to-end manner.

\setlength{\tabcolsep}{0.5pt} 
\renewcommand{\arraystretch}{1.1} 
\begin{table}[t]
    \normalsize
    \centering
    \caption{\textbf{Analysis of AuxDepthNet components on the KITTI test set (Car category).}The table presents the performance (AP$_{3D}$ @ IoU=0.7) of various configurations, including the baseline, module variations, and the full AuxDepthNet model.}
    \label{tab:a1}
    \begin{tabular}{|c|c|c|c|c|}  
        \hline  
        \multirow{2}{*}{\centering } & 
        \multirow{2}{*}{\centering \textbf{Ablation}} & 
        \multicolumn{3}{c|}{\textbf{AP$_{3D}$@IoU=0.7}} \\ \cline{3-5} 
        & & \textbf{Easy} & \textbf{Mod.} & \textbf{Hard} \\ 
        \hline
        (a) & Baseline & 19.50 & 15.51 & 12.66 \\
        (b) & w/o depth prototype enhancement & 23.91 & 18.27 & 15.21 \\
        (c) & Depth-sensitive feature $\rightarrow$ object query & 20.25 & 16.15 & 13.89 \\
        (d) & Depth-sensitive feature $\rightarrow$ DORN \cite{FuCVPR18-DORN} & 24.27 & 17.15 & 13.84 \\
        (e) & AuxDepthNet (full model) & \textbf{24.72} & \textbf{18.63} & \textbf{15.31} \\ 
        \hline
    \end{tabular}
    \vspace{-2mm}
\end{table}

The results in Table \ref{tab:a1} highlight several key observations: substituting object queries in the transformer with depth-sensitive features (c$\rightarrow$e) significantly enhances depth representation, resulting in notable performance improvements. Furthermore, using features derived from a pretrained depth estimator (d) underperforms compared to the end-to-end approach in our full framework (e), emphasizing the benefits of directly learning depth-sensitive features.
Our depth-sensitive transformer module (e) effectively combines context- and depth-sensitive features, outperforming the simpler convolutional concatenation method (b). Additionally, the inclusion of the depth prototype enhancement module (d$\rightarrow$e) further refines detection accuracy, particularly under more challenging scenarios.Ultimately, the complete AuxDepthNet model (e) delivers marked advancements across all levels of difficulty relative to the baseline (a), demonstrating the combined contributions of the proposed modules to robust and precise 3D object detection.

\subsubsection{The impact of different backbones.}
\setlength{\tabcolsep}{3pt} 
\renewcommand{\arraystretch}{1.2} 
\begin{table}[t]
    \caption{\textbf{Comparison of Different Backbones on KITTI Test Set for Car Category}The table presents the performance (AP$_{3D}$ and AP$_{\text{BEV}}$ @ IoU=0.7) using various backbones, including our proposed AuxDepthNet.}
    \label{tab:a2}
    \centering
    \normalsize 
    \begin{tabular}{|>{\centering\arraybackslash}m{2cm}|c|c|c|c|c|c|} 
        \hline   
        \multirow{2}{*}{\textbf{Backbone}} & \multicolumn{3}{c|}{\textbf{AP$_{3D}$@IoU=0.7}} & \multicolumn{3}{c|}{\textbf{AP$_{\text{BEV}}$@IoU=0.7}} \\ 
        \cline{2-7}
        & \textbf{Easy} & \textbf{Mod.} & \textbf{Hard} & \textbf{Easy} & \textbf{Mod.} & \textbf{Hard} \\ 
        \hline
        DLA-102            & 24.40 & 18.55 & 15.30 & 34.02 & 25.11 & 21.24 \\ 
        DLA-102X2          & 24.59 & 18.56 & 15.25 & 34.01 & 25.13 & 21.32 \\ 
        DLA-60             & 22.85 & 17.37 & 14.33 & 32.14 & 23.64 & 19.92 \\ 
        DenseNet           & 24.24 & 18.04 & 15.14 & 33.01 & 24.69 & 21.48 \\ 
        ResNet             & 24.09 & 17.98 & 15.15 & 32.97 & 24.39 & 21.33 \\ 
        \textbf{Ours} & \textbf{24.72} & \textbf{18.63} & \textbf{15.31} & \textbf{34.11} & \textbf{25.18} & \textbf{21.90} \\ 
        \hline
    \end{tabular}
\end{table}
To further evaluate the impact of different backbone networks on the performance of 3D object detection for the Car category in the KITTI test set, we conducted an ablation study comparing various backbones. Specifically, we tested DLA-102, DLA-102X2, DLA-60, DenseNet, ResNet, and our proposed AuxDepthNet (Ours). Among these, our framework employs DLA-102 as the default backbone due to its balanced trade-off between efficiency and accuracy. The\begin{figure*}[htbp]
  \centering
  \includegraphics[width=1\textwidth]{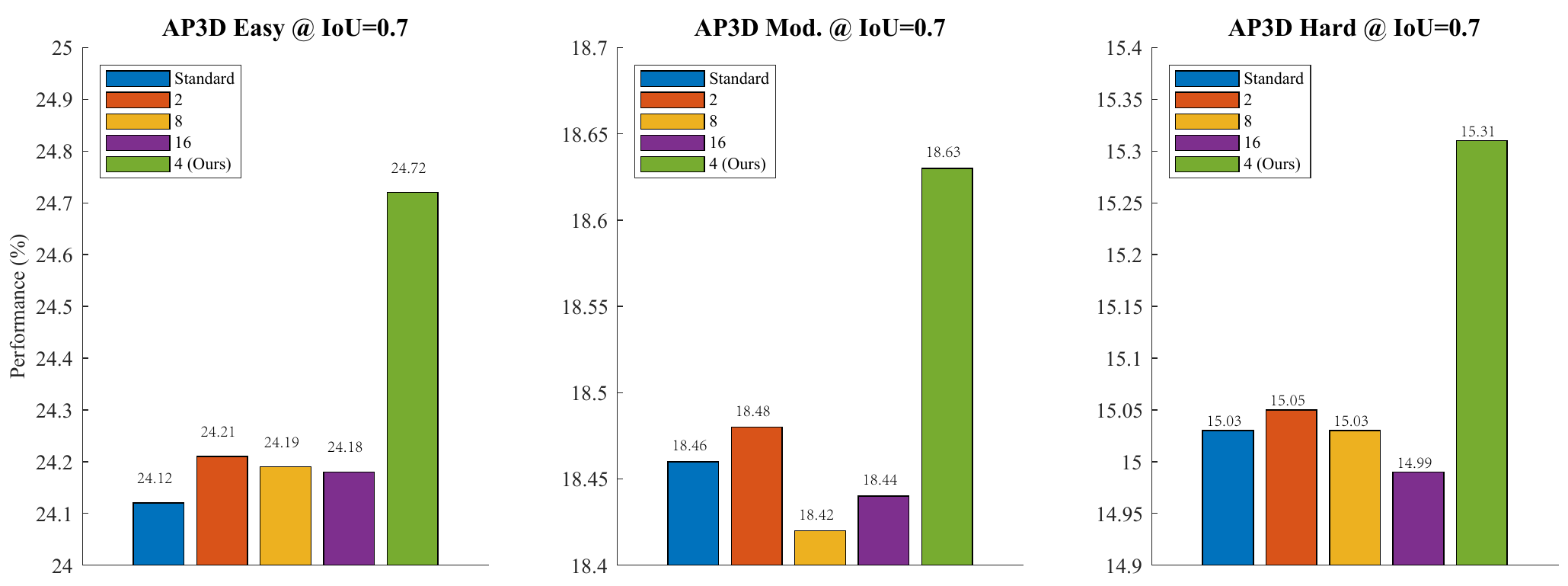}
  \caption{\textbf{Comparison of AP$_{\text{3D}}$ detection accuracy for the Car category on the KITTI validation set using different dilation rates in the dilated convolution of the ADF module.}Comparison of AP$_{\text{3D}}$ detection accuracy for the Car category on the KITTI validation set, with standard convolution replaced by dilated convolutions using dilation rates of 2, 4, 8, and 16 in the ADF module.
  }
  \label{fig:pdf-fixed}
\end{figure*}\begin{figure*}[htbp]
    \centering
    \includegraphics[width=0.95\textwidth]{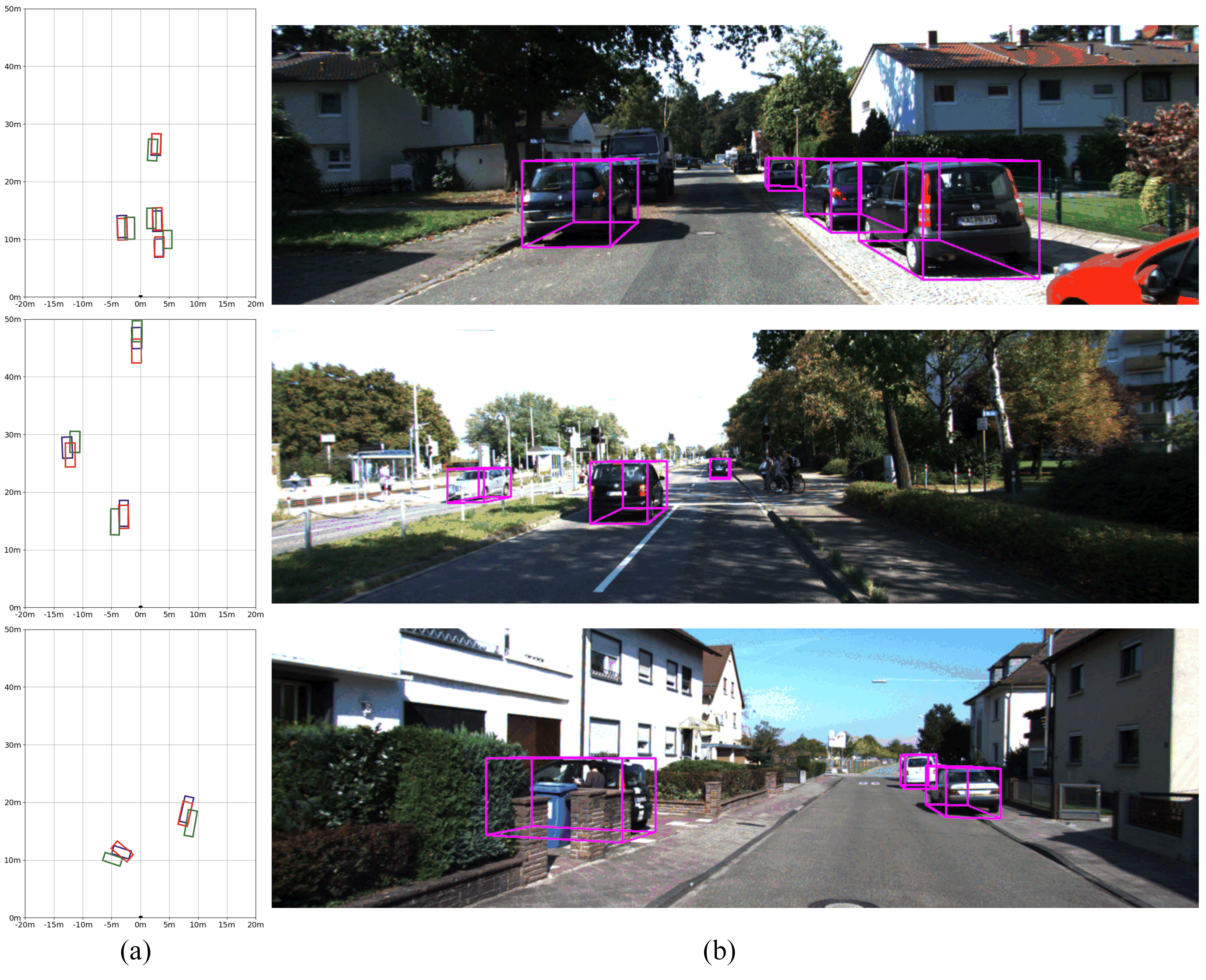}
    \caption{\textbf{Qualitative results on the KITTI dataset.} (a) Bird’s-eye view of detection results, where blue represents the ground truth (GT), red denotes predictions from the AuxDepthNet model, and green indicates baseline (without depth-sensitive modules) predictions. (b) Predicted results from the AuxDepthNet model visualized in the image view. (Zoom in for a detailed view.)}
    \label{fig:wide_image}
\end{figure*} results demonstrate that while other backbones provide competitive performance, DLA-102 achieves superior precision, validating its suitability for our approach. The result show in table \ref{tab:a2}.

The ablation study highlights the impact of different backbones on KITTI’s Car category detection. DLA-102 achieves a strong balance between efficiency and accuracy, while DLA-102X2 slightly outperforms it, particularly in the Easy and Moderate settings. DenseNet and ResNet provide competitive results but fall short of DLA-based models, and DLA-60 demonstrates limited performance due to reduced capacity. Notably, our proposed AuxDepthNet achieves the highest accuracy, with AP$_{\text{3D}}$ of 18.63 (Moderate) and APBEV of 25.18 (Moderate), showcasing its superior depth-sensitive feature extraction capabilities for monocular 3D detection.So our AuxDepthNet backbone outperforms conventional backbones, highlighting the effectiveness of its design in enhancing depth-sensitive feature representation.

\subsubsection{The impact of different dilation rates in the ADF module.}
To investigate how various dilation rates in the ADF module affect 3D object detection performance, we conducted ablation experiments on the validation set of the KITTI dataset, focusing on the Car category. Specifically, we set the dilation rates to Standard (no dilated convolution), 2, 8, 16, and 4 (Ours), all under the same network architecture and training hyperparameters, and evaluated their $AP_{3D}$ (Average Precision in 3D) metrics. As shown in Fig.~\ref{fig:pdf-fixed}, different dilation rate configurations yield varying results across the Easy, Moderate, and Hard difficulty levels. Notably, a dilation rate of 4 achieves the highest detection accuracy, reaching 24.72\%, 18.63\%, and 15.31\% on Easy, Moderate, and Hard, respectively---outperforming the other settings. These findings suggest that using a suitable dilation rate helps capture more effective multi-scale features, thereby significantly enhancing the accuracy of 3D object detection. Hence, we choose a dilation rate of 4 for all subsequent experiments to achieve optimal performance.

\section{Visualization}
Fig.\ref{fig:wide_image} presents qualitative examples from the KITTI validation set. Compared to the baseline model without depth-sensitive modules, AuxDepthNet predictions align much more closely with the ground truth, demonstrating the effectiveness of the proposed depth-sensitive modules in enhancing object localization accuracy.
\section{Conclusion}\label{sec:con} 
In this study, we proposed AuxDepthNet, a framework for real-time monocular 3D object detection that eliminates the need for external depth maps or pre-trained depth models. By introducing the Auxiliary Depth Feature Module (ADF) and the Depth Position Mapping Module (DPM), AuxDepthNet effectively learned depth-sensitive features and integrated depth positional information, enhancing spatial reasoning with minimal computational cost. Built on the DepthFusion Transformer architecture, the framework demonstrated robust performance in object localization and 3D bounding box regression.

Despite its effectiveness, AuxDepthNet has limitations, including its focus on the KITTI dataset and potential challenges in generalizing to diverse environments. Future work will explore broader dataset adaptation, improve robustness under extreme conditions, and optimize computational efficiency for edge applications.

\section*{Acknowledgment}
Authors appreciate the support of family and friends who encouraged and inspired them during the course of this work.

\bibliographystyle{IEEEtran}
\bibliography{ref}

\EOD

\end{document}